\pdfoutput=1

\documentclass[11pt]{article}

\usepackage[final]{acl}

\usepackage{times}
\usepackage{latexsym}

\usepackage[T1]{fontenc}

\usepackage[utf8]{inputenc}

\usepackage{microtype}

\usepackage{inconsolata}

\usepackage{graphicx}

\usepackage{booktabs}
\usepackage{float}
\usepackage{multirow}
\usepackage{tabularx}
\usepackage{makecell}

\title{Towards a Principled Evaluation of Knowledge Editors}

\author{Sebastian Pohl \and Max Ploner \and Alan Akbik
  \vspace{0.6em} \\
  Humboldt Universität zu Berlin \\
  Science Of Intelligence \vspace{.2em}\\
  \texttt{<first name>.<last name>@hu-berlin.de} \\
}

\begin{document}
\maketitle

\newcommand{\promptAnswer}[2]{{
    \setlength\fboxsep{.8em}
    \fbox{%
        \parbox{0.9\linewidth}{%
            \footnotesize
            \setlength{\parskip}{.5em}
            \textit{#1}\\
            \rule[0.25em]{\linewidth}{0.4pt}
            #2
        }
    }
}}

\let\originalparagraph\paragraph
\renewcommand{\paragraph}[2][.]{\originalparagraph{#2#1}}

\begin{abstract}

Model editing has been gaining increasing attention over the past few years. For Knowledge Editing in particular, more challenging evaluation datasets have recently been released. These datasets use different methodologies to score the success of editors. Yet, it remains under-explored how robust these methodologies are and whether they unfairly favor some editors. Moreover, the disruptive impact of these editors on overall model capabilities remains a constant blind spot.

We address both of these problems and show that choosing different metrics and evaluation methodologies as well as different edit batch sizes can lead to a different ranking of knowledge editors. Crucially we demonstrate this effect also on general language understanding tasks evaluated alongside the knowledge editing tasks. Further we include a manual assessment of the string matching based evaluation method for knowledge editing that is favored by recently released datasets, revealing a tendency to produce false positive matches.

\end{abstract}

\section{Introduction}
Pre-trained language models have been demonstrated to perform well on a wide variety of NLP tasks and applications even without the need for specific fine-tuning \cite{brown2020languagemodelsfewshotlearners}. Nonetheless, researchers have sought to adjust models to their specific needs even outside of the fine-tuning paradigm. Continual Learning focuses on the need to update models beyond their training cutoff date or to adapt them to new domains without the need for full re-training \cite{Kirkpatrick_2017, Biesialska_2020}. Retrieval-Augmented Generation (RAG) is being used to improve performance on domain-specific or knowledge-intensive tasks or to reduce the number of ``hallucinations'' language models produce \cite{lewis2021retrievalaugmentedgenerationknowledgeintensivenlp, gao2024retrievalaugmentedgenerationlargelanguage}.

Building on these techniques, Model Editing has emerged as an independent research direction. In principle, Model Editing is agnostic to the specific method used to adjust model behavior. It defines targeted local changes to the desired model outputs, such as correcting specific errors, updating individual pieces of knowledge or the sentiment towards specific entities \cite{sinitsin2020editableneuralnetworks, pmlr-v162-mitchell22a, ilharco2023editingmodelstaskarithmetic}.
The techniques used to effectuate these desired changes include the training of  hyper-networks \cite{decao2021editingfactualknowledgelanguage}, explicitly calculated parameter updates \cite{meng2023locatingeditingfactualassociations, meng2023masseditingmemorytransformer}, and in-context learning \cite{zheng2023editfactualknowledgeincontext, cohen2023evaluatingrippleeffectsknowledge}. The latter is closely related to RAG since in in-context learning, natural language expressions of the knowledge are prepended to the model prompts. These injected sentences may, in turn, be retrieved from some external knowledge store.

Knowledge Editing, where new knowledge (often given by relation triplets $\langle$\textit{subject, relation, object}$\rangle$) is injected into the model, is the most common but not the only variant of Model Editing. Any targeted and local updates to model behavior could be subsumed under Model Editing, including, for example also such topics as unlearning, where specific pieces of private or harmful information should not be produced by the model \cite{jang2022knowledgeunlearningmitigatingprivacy, hong2024intrinsicevaluationunlearningusing}.

\paragraph{Research Gaps and our Objectives}
Our experiments are primarily focused on Knowledge Editing. Previous work has established four datasets for the evaluation of knowledge editors: \textit{zsre} \cite{levy-etal-2017-zero}, \textit{CounterFact} \cite{meng2023locatingeditingfactualassociations}, \textit{MQuAKE} \cite{zhong2024mquakeassessingknowledgeediting}, and \textit{RippleEdits} \cite{cohen2023evaluatingrippleeffectsknowledge}.
These datasets include different types of queries to test for the efficacy and locality of edits as well as the ability of models to draw inferences from edited knowledge.
But they also use different methods and metrics to score whether edited models are successful at effecting the desired edits.
Specifically, \textit{zsre} classifies token by token, whether greedy decoding produces the desired output, \textit{CounterFact} uses a ranking of alternative answers by sequence log likelihoods, and \textit{MQuAKE} and \textit{RippleEdits} test if target strings match within text generated in answer to query prompts.
So far, the impact of these evaluation methods has not been analysed. Our experiments show that one of the editors we tested, \textit{MEMIT} \cite{meng2023masseditingmemorytransformer}, does better than other editors, specifically when it is evaluated based on the ranking of alternative sequence log likelihoods. While the evaluation by matching expected answers in generated query responses is favored by more recently released datasets, the validity of this method and where it fails also remains under-explored.

Secondly, it seems evident that the more edits an editor has to inject into a model, the more difficult this task becomes and the more disruptive the editing is for the overall model performance. While some editors are designed specifically to inject a large number of edits, including the aforementioned \textit{MEMIT} editor, the relationship between the number of edits and the changes in model performance deserve a more systematic study. In particular, where it concerns not only the Knowledge Editing performance, but also the retention of overall model capabilities. We demonstrate this gap by evaluating editors on a wide range of edit batch sizes and by integrating Knowledge Editing datasets with LM Evaluation Harness \cite{eval-harness} to run general language understanding tasks on edited models alongside the Knowledge Editing evaluation.

\paragraph{Contributions}
In this study, we aim to demonstrate the influence of possible design choices on the outcomes of knowledge editing benchmarks (and evaluation setups).
We focus on making the effects of these choices more explicit over evaluating the exact ranking of specific Model Editing methods. We hope our findings may support more informed evaluation practices and encourage further research in this area.

Together with our research, we also release the evaluation framework used in our experiments as open source. It combines the four mentioned existing Knowledge Editing datasets into a unified framework, integrates with LM Evaluation Harness to allow for the evaluation of edited models on its tasks, and can easily be extended with support for additional models, evaluation datasets, and model editors.\footnote{Our framework is available at \href{https://github.com/oneSebastian/model-editing}{model editing}. The evaluation results of our experiments and code used for figures in this paper are available at \href{https://github.com/oneSebastian/model-editing-results}{paper results}.}

\section{Background}
While the framework we use for our experiments can be used for different types of Model Editing, our experiments are focused on the evaluation of Knowledge Editing methods.

Exact formalisms vary, but generally, Knowledge Editing is defined along the following lines:

Assuming a model $x\mapsto f(x,\theta)$ with trained parameters $\theta$, we are given a set of revisions $\langle x, y, a\rangle\in D$, where $x$ is some model input, $y$ is the output preferred by $f(x,\theta)$ and $a$ is the post-edit output we would like the model to prefer instead. Additionally, for evaluation, a revision $\langle x, X, y, a\rangle$ may contain a set $X$ of inputs $x', x'', ...$ that are semantically equivalent to $x$ \cite{decao2021editingfactualknowledgelanguage, sinitsin2020editableneuralnetworks}. Knowledge Editors were originally evaluated on three metrics \cite{mitchell2022fastmodeleditingscale}. Assuming an edit $\langle x_1, X_1, y_1, a_1\rangle\in D$:

\textbf{Reliability:} the post edit model predicts the output $a_1$ given input $x_1$.

\textbf{Locality:} for unrelated entries $\langle x_i, y_i, a_i\rangle\in D$ the model continues to predict $y_i$ given input $x_i$.

\textbf{Generalizability:} the model also predicts $a_1$ given a semantically equivalent input $x_1'\in X_1$.

\subsection{Datasets}\label{sec:datasets}
Two datasets are primarily used for evaluation along these metrics: \textit{zsre} \cite{levy-etal-2017-zero} and \textit{CounterFact} \cite{meng2023locatingeditingfactualassociations}. They both consist of entries that contain some edit fact \textit{(subject, relation, object)} expressed through a natural language template together with a number of queries that test for reliability, locality, and generalizability (see appendix \ref{sec:appendix} for examples from each dataset). \textit{CounterFact} was introduced alongside \textit{zsre}, as the latter proved insufficiently challenging. Unedited models often already assign high scores to the correct edit outputs. This is avoided by using counterfactual edits, where the post-edit target would not have been part of the model's training data \cite{meng2023locatingeditingfactualassociations}.

Researches have also measured the generation quality of the post-edit models by scoring the TF-IDF similarity between text generated by an edited model given a prompt such as \textit{``Michael Jordan plays the sport of''} and a Wikipedia reference article about the target object \textit{``basketball''} as well as scoring the entropy of bi- and tri-gram n-gram distributions of generated text \cite{meng2023locatingeditingfactualassociations, meng2023masseditingmemorytransformer}.

These two datasets test whether edited models can recall edit facts while unrelated facts remain unchanged. More recently, additional Knowledge Editing benchmarks have been introduced that cover abilities an edited model should possess unaddressed by \textit{zsre} or \textit{CounterFact}. \textit{MQuAKE} covers the question of whether edited knowledge is utilized in multi-hop reasoning \cite{zhong2024mquakeassessingknowledgeediting}. If, for example, we insert the new knowledge that \textit{``Keir Starmer is the Prime minister of the UK.''} instead of his predecessor \textit{``Rishi Sunak''}, the post-edit model should also produce an updated answer to the question \textit{``Who is the spouse of the British prime minister?''} or any other implied facts. \textit{RippleEdits}, another more recent benchmark, also contains some test cases for multi-hop reasoning as well as other types of inferences based on properties of the relations present in edit triplets and queries to test if edited models forget knowledge the pre-edit model possessed \cite{cohen2023evaluatingrippleeffectsknowledge}.

\subsection{Model Editors}
To keep the number of required experiments at a manageable level, we only included a select number of model editors. Our study focuses on the evaluation of these model editors. For wider surveys of larger ranges of editors and editing methods, see, for example, \cite{yao2023editinglargelanguagemodels, zhang2024comprehensivestudyknowledgeediting}.

First, we included \textit{MEMIT} \cite{meng2023masseditingmemorytransformer} as one of the most promising variants of editors that update model parameters. It is designed specifically to inject a large amount of edits and is widely used as a well-performing baseline in related work. \textit{Memit} calculates explicit parameter updates through causal tracing based on gradients to inject individual edits into specific model layers.

Second, we include \textit{LoRA}, an editor based on the popular LoRA technique \cite{hu2021loralowrankadaptationlarge} and used as a Knowledge Editing baseline for example in \cite{zhang2024comprehensivestudyknowledgeediting}. We consider full fine-tuning on individual edits to be too resource intensive, and include this variant of parameter efficient fine tuning as an alternative instead.

Third, we included a simple \textit{in-context} editor, that has been shown to be particularly effective for more challenging recent knowledge datasets \cite{zheng2023editfactualknowledgeincontext, cohen2023evaluatingrippleeffectsknowledge}. The \textit{in-context} editor just prepends edit facts expressed through natural language templates to the model inputs and leaves the integration up to the model's attention mechanism.

Fourth, we implement a \textit{context-retriever} editor that also just prepends edits to model inputs. However, with the size of the context window, there is a clear limit to how many edits such an editor can inject. We, therefore, combine the \textit{in-context} editor with a RAG system. We follow \cite{zhong2024mquakeassessingknowledgeediting} in using the Contreiver model \cite{izacard2022unsuperviseddenseinformationretrieval} to encode all edits and retrieve 4-NN edits given any query. We chose 4-NNs, because the MQuAKE examples depend at most on four edits for 4-hop reasoning. Unlike \cite{zhong2024mquakeassessingknowledgeediting}, however, we do not include any chain of thought reasoning, such as generating sub-questions and answering them separately to improve multi-hop reasoning. A basic tenet of our inquiry is that an edited model should behave just like a normal language model that immediately generates text in response to an input prompt.

As a baseline, we also include results for an unedited model. In our experiments, we do batch model editing, where an editor has to inject $n$ edits simultaneously for an edit batch size of $n$.

\section{Scoring and Metrics}
The datasets mentioned in section \ref{sec:datasets} do not only use different types of test queries to test for reliability, locality, generalizability, multi-hop reasoning, and other types of inferences, they also use different methods to score whether a model produces the correct post-edit output.

\textbf{Argmax:} In \textit{zsRE}, each test case comes with a prompt and a desired target string. In evaluation, it is then tested, token by token, whether each token of the target string is assigned the highest probability, i.e., if it would be produced by greedy decoding. The score for the test case is the average over this binary decision, i.e., an accuracy score of $0.75$, if $3$ out of $4$ target tokens are assigned maximum logits.

\textbf{MC:} In \textit{CounterFact}, each test case prompt has an original and a new post-edit target because each edit fact is counterfactual, replacing a true target by a supposed new edit target. Test cases are then treated as a \textit{multiple choice} task. The likelihood of the entire sequence is scored both with the original and the new target and a test case is counted as a success if the new target sequence is scored as more likely by the edited language model.

\textbf{Generate:} \textit{MQuAKE} and \textit{RippleEdits} provide original as well as new targets only for edit facts but not for test cases. Instead, each fact and test case also has a number of aliases for the post edit target. Test cases are scored by generating a fixed number of tokens for each test case prompt and by checking if any of the new target aliases are contained in the generated text.

Firstly, while \textit{argmax} and \textit{generate} can be used with all four datasets, only CounterFact consistently contains the answer alternatives required for a \textit{multiple choice} evaluation. However, so far, it remains unclear if these different evaluation methods produce the same results or if it would be feasible to use the same method for all datasets.

Secondly, with the length of the generated text, the \textit{generate} method includes a critical hyper-parameter that has to be tuned appropriately. Conceivably, in some cases, a model may require more tokens to produce an answer containing the target string. But equally, a longer generated answer may increase the rate of false positive matches.

\subsection{Experimental Setup}\label{sec:experimental_setup}
In our experiments, we want to address both of these questions. To answer the first question, comparing \textit{argmax}, \textit{multiple choice} and \textit{generate}, we evaluated the four knowledge editors \textit{MEMIT}, \textit{LoRA}, \textit{in-context} and \textit{context-retriever} on all included Knowledge Editing datasets using every scoring methods applicable to the given dataset. RippleEdits has a total number of 4655 viable examples, zsRE has 19086, MQuAKE 3000, and CounterFact 21919 examples. To save compute we randomly selected 2048 examples from each dataset for all our experiments, drawing evenly from each dataset split in the case of RippleEdits.

We ran these and all later experiments on two models, GPT-J with 6B parameters \cite{gpt-j} and GPT2-xl with 1.5B parameters \cite{gpt2-xl}. These models were chosen because they are also used in the related literature and because the authors of MEMIT have published hyper-parameters needed for their editor only for these two models \cite{meng2023masseditingmemorytransformer}. For \textit{LoRA} we briefly explored a range of hyperparameters optimizing for performance on an edit batch size of 16. We observed that smaller batch sizes generally benefited from higher learning rates, likely because the adapter needs to be fitted on fewer examples and thus fewer optimization steps. Based on these findings, we used the following \textit{LoRA} hyperparameters in our experiments: a rank of 8, an alpha value of 32, and 20 training epochs (i.e., 20 passes over the edit batch). For GPT-2-XL, we used a learning rate of 5e-3, and for GPT-J, 1e-3.

To address the second question of how much text to generate in response to a query prompt, we evaluated all editors on all Knowledge Editing datasets with the \textit{generate} method, generating 64 tokens of text given a query prompt. We then calculated accuracy scores for any generation length up to 64 tokens.

For 200 of these examples, we also manually evaluated the quality of the exact string matching-based evaluation method. We first separated examples depending on whether at least one of the editors achieved a \textit{late success}, i.e., produced a matching answer in the second half of the generated text, but not earlier. We drew an equal number of examples from each dataset for both this \textit{late success} class and its complement, the \textit{early success} class. Examples in the \textit{early success} class were either immediately answered correctly or not answered correctly at all by all editors (as can be seen in figure~\ref{fig:no_late_success_stats} in the appendix). Since we were interested in the effect of generating longer stretches of text, we focused on the \textit{late success} class and evaluated 150 examples from this and 50 examples from the \textit{early success} class.

Raters were given the responses generated by edited models for query prompt and the post-edit expected answers. They were asked to judge whether the first answer given by the model correctly answers the prompt.

\subsection{Results}\label{sec:results_scoring_and_metrics}
Tables~\ref{tab:gptj_evaluation_method} and~\ref{tab:gpt2_xlevaluation_method} show the accuracy results for all datasets, editors, and compatible evaluation methods for GPT-J and GPT2-XL, respectively. These experiments were conducted with an edit batch size of 16. When \textit{generate} was used, models produced 20 tokens in response to any given prompt.

\begin{table}[ht]
    \centering
    \resizebox{\columnwidth}{!}{
        \begin{tabular}{c|c|ccccc}
            \toprule
            \textbf{Dataset} & \textbf{Eval} & \textbf{CR} & \textbf{IC} & \textbf{MEMIT} & \textbf{LoRA} & \textbf{NoEdit} \\
            \midrule
            \multirow{2}{*}{zsRE} & argmax & 0.735 & \textbf{0.764} & 0.727 & 0.756 & 0.278 \\
                                          & gen & 0.619 & \textbf{0.656} & 0.629 & 0.653 & 0.066 \\
            \midrule
            \multirow{3}{*}{CF} & argmax & \textbf{0.365} & 0.391 & 0.312 & 0.356 & 0.095 \\
                                          & MC & 0.800 & 0.794 & \textbf{0.866} & 0.688 & 0.614 \\
                                          & gen & 0.505 & \textbf{0.511} & 0.462 & 0.442 & 0.200 \\
            \midrule
            \multirow{2}{*}{MQuAKE} & argmax & 0.330 & \textbf{0.345} & 0.211 & 0.300 & 0.204 \\
                                          & gen & \textbf{0.213} & 0.198 & 0.153 & 0.133 & 0.050 \\
            \midrule
            \multirow{2}{*}{RipEd} & argmax & 0.621 & \textbf{0.626} & 0.502 & 0.591 & 0.353 \\
                                          & gen & 0.500 & 0.478 & 0.475 & 0.537 & \textbf{0.543} \\
            \bottomrule
        \end{tabular}
    }
    \caption{Accuracy scores for different evaluation methods on GPT-J.}
    \label{tab:gptj_evaluation_method}
\end{table}
\begin{table}[ht]
    \vspace{-3mm}
    \centering
    \resizebox{\columnwidth}{!}{
        \begin{tabular}{c|c|ccccc}
            \toprule
            \textbf{Dataset} & \textbf{Eval} & \textbf{CR} & \textbf{IC} & \textbf{MEMIT} & \textbf{LoRA} & \textbf{NoEdit} \\ 
            \midrule
            \multirow{2}{*}{zsRE} & argmax & 0.718 & \textbf{0.724} & 0.495 & 0.595 & 0.239 \\
                                          & gen & 0.604 & \textbf{0.619} & 0.322 & 0.542 & 0.049 \\
            \midrule
            \multirow{3}{*}{CF} & argmax & \textbf{0.330} & 0.310 & 0.205 & 0.234 & 0.072 \\
                                          & MC & 0.766 & 0.745 & \textbf{0.779} & 0.680 & 0.596 \\
                                          & gen & \textbf{0.444} & 0.404 & 0.313 & 0.291 & 0.135 \\
            \midrule
            \multirow{2}{*}{MQuAKE} & argmax & 0.318 & \textbf{0.334} & 0.190 & 0.081 & 0.189 \\
                                          & gen & \textbf{0.325} & 0.208 & 0.085 & 0.076 & 0.060 \\
            \midrule
            \multirow{2}{*}{RipEd} & argmax & \textbf{0.632} & 0.594 & 0.487 & 0.377 & 0.374 \\
                                          & gen & 0.542 & 0.433 & 0.499 & 0.391 & \textbf{0.562} \\
            \bottomrule
        \end{tabular}
    }
    \caption{Accuracy scores for different evaluation methods on GPT2-XL.}
    \label{tab:gpt2_xlevaluation_method}
    \vspace{-5mm}
\end{table}

We observe that while, for the most part, all evaluation methods produce the same relative ranking of model editors, there are a few notable exceptions. On the CounterFact dataset, \textit{MEMIT} outperforms the other editors according to the \textit{multiple choice} evaluation method that is the authors' choice for this dataset \cite{meng2023locatingeditingfactualassociations} but performs worst according to both other methods. On MQuAKE, \textit{in-context} outperforms \textit{context-retriever} according to the \textit{argmax} method, but this is reversed with the dataset default \textit{generate} method. Despite performing worse overall \textit{LoRA} out performes \textit{MEMIT} on some datasets and on \textit{GPT-J} all other editors on the \textit{RippleEdits} dataset, when evaluated with the \textit{generate} method. Unlike for other editors, however, we specifically tuned the \textit{LoRA} hyper-parameters to the edit batch size of 16. As can be seen in section \ref{sec:evaluation_settings_results} this comes at a price for other edit batch sizes and general language understanding tasks.

Next to these order reversals, the absolute differences also vary. While \textit{MEMIT} barely performs better than an unedited model on MQuAKE evaluated with the \textit{argmax} method, its accuracy is three times as high on \textit{GPT-J} according to the \textit{generate} method. Overall, accuracy results are not very robust between these alternative evaluation methods.

\begin{figure}[ht]
    \vspace{-8mm}
    \centering
    \includegraphics[width=\columnwidth]{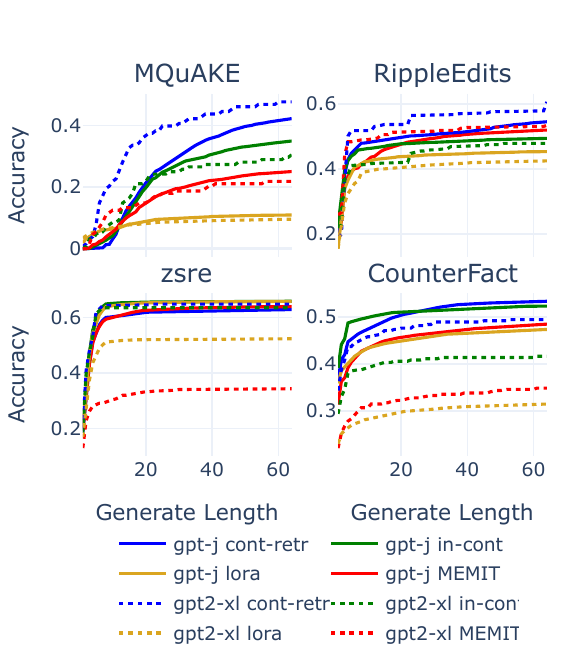}
    \vspace{-6mm}
    \caption{Accuracies for different Model Editors and datasets on different numbers of generated tokens.}\label{fig:generate_length}
    \vspace{-1mm}
\end{figure}

Figure~\ref{fig:generate_length} shows the accuracy scores over the four benchmark datasets for varying lengths of generated text (counted in number of generated tokens). Experiments were run with an edit batch size of 16 over 2048 examples from each dataset. Particularly on \textit{zsre}, all models achieve their final accuracy after a short number of generated tokens already, i.e., if the edited model is not immediately generating an accepted answer, it will not generate one at all. There is an interesting difference in the relative ranking of the editors. On GPT-J, the \textit{context retriever} benefits from an increase in the number of generated tokens relative to the \textit{in-context} editor. While on GPT2-XL, the former outperforms the latter already on shorter generated answers. We address a possible cause for this in our manual evaluation of model answers.

\begin{figure*}[ht]
    \centering
    \includegraphics[width=\textwidth]{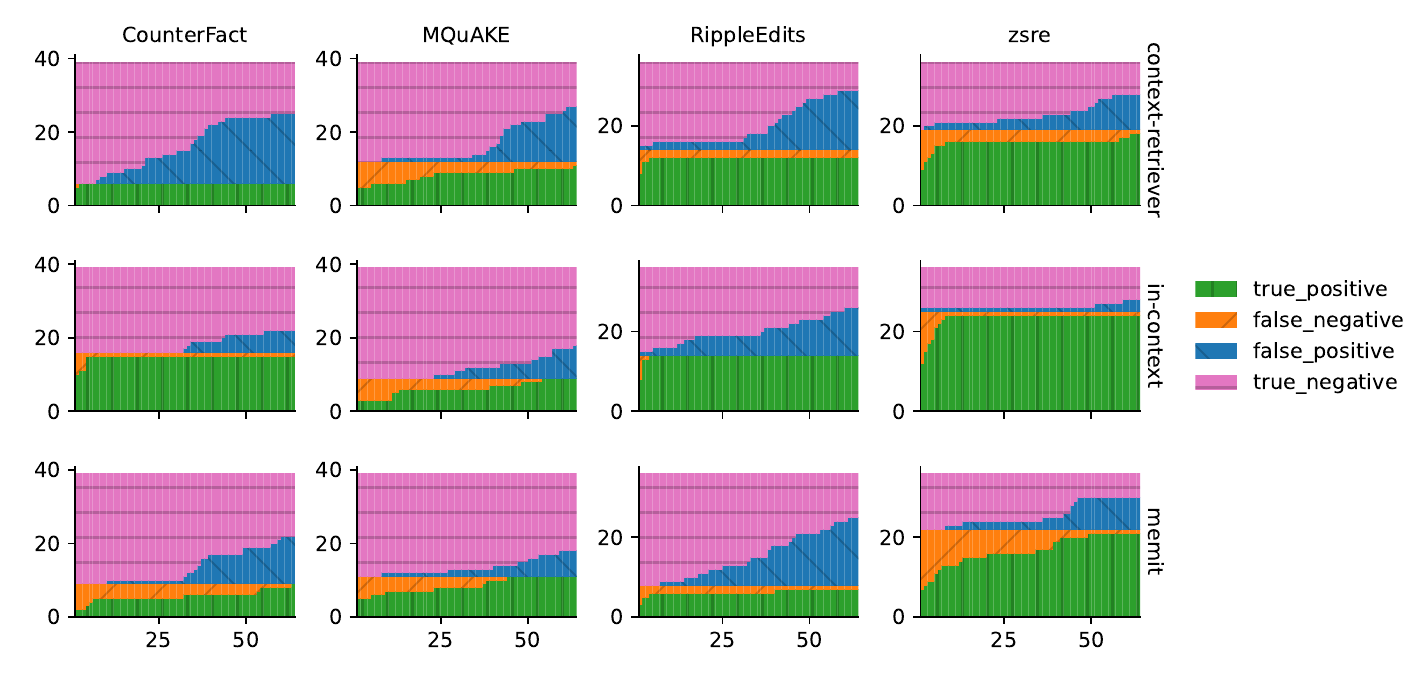}
    \caption{True Positives, True Negatives, False Positives, and False Negatives for each Editor, Dataset, and Generate Length on GPT-J (only including samples where at least one editor generates a first correct substring in the second half of the answer).}
    \label{fig:late_success_stats}
\end{figure*}

Out of the 200 examples we manually rated, 150 belong to the \textit{late success} class, where at least one of the editors generated a correct answer in the second half of the generated text and not earlier. For these examples, Figure~\ref{fig:late_success_stats} plots the true positives, true negatives, false positives, and false negatives for each editor, dataset, and generate-length on GPT-J against each other. We just observed that the \textit{context-retriever} benefits from longer generation lengths compared to \textit{in-context}. In this figure, we can see that that is likely due to a larger false positive rate for the \textit{context-retriever} as the length of generated text increases.

\begin{figure}[ht]
    \vspace{-2mm}
    \centering
    \includegraphics[width=\columnwidth]{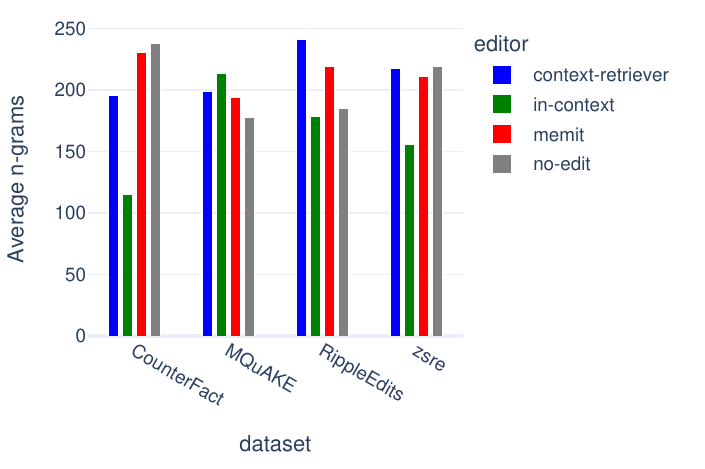}
    \vspace{-8mm}
    \caption{Average number of unique $n\leq 5$-grams per generated 64 answer tokens for different datasets and editors.}\label{fig:n_grams}
\end{figure}

We speculated that a model edited with the \textit{context-retriever} generates more varied text, resulting in a higher chance to produce false positives. Figure~\ref{fig:n_grams} shows the average number of unique token $n\leq 5$-grams for the answers generated in the \textit{late success} examples. While for most datasets the \textit{context-retriever} model did produce more varied text than its \textit{in-context} counterpart, this is reversed for MQuAKE, even though on this dataset we also observed that the \textit{context-retriever} accuracy score exceeds the \textit{in-context} accuracy score due to a higher false positive rate. The exact cause of the difference between \textit{in-context} and \textit{context-retriever} remains unclear.

Despite the naïve matching scheme (exact sub-strings), the rate of false negatives is relatively small (assuming at least 10 tokens are generated). At least partially, this may be due to the relatively aggressive matching. For example, Figure~\ref{fig:example_answer_rome} shows a case where a close answer is given, and the answer is considered correct, though not for the right reason. The fact that the Italian Parliament is located in Rome here does not matter. The exact match is found in an unrelated piece of information (the place of birth).

\begin{figure}
    \promptAnswer{Alcide De Gasperi worked in}{
        the Italian Parliament for over 30 years. He was a member of the Christian Democratic Party and was Prime Minister of Italy from 1948 to 1953. He was also President of the European Parliament from 1958 to 1959.

        Alcide De Gasperi was born in \textbf{Rome} on April 26, 1881. He was the
    }
    \caption{Prompt (in \textit{italcis}) and generated answer (matched substring marked in \textbf{bold}).}
    \label{fig:example_answer_rome}
\end{figure}

\begin{figure}
    \promptAnswer{what is the main mineral in lithium batteries?}{
        A:
        
        Lithium is the main component of the anode. The cathode is made of carbon and the electrolyte is a mixture of \textbf{lithium} salts and organic solvents.
        
        A:
        
        Lithium is the main component of the anode. The cathode is made
    }
    \caption{Prompt (in \textit{italcis}) and generated answer (matched substring marked in \textbf{bold}).}
    \label{fig:example_answer_lithium}
\end{figure}

The example in Figure~\ref{fig:example_answer_lithium} is even more striking: The initial answer (\textit{``Lithium''}) may be considered correct but is ignored by the exact matching algorithm since it is capitalized, but the expected answer is not. Only later is the answer deemed as correct due to another match.
One might consider using case-insensitive matching. However, we believe that while this would solve this particular issue, it would introduce more false negatives. A more sophisticated matching algorithm, however, may help in avoiding these issues.

As an additional alternative we also tried an LLM-as-a-judge approach. Instruction tuned models (\textit{Mistral-7B-Instruct-v0.3} \cite{mistral7b} and \textit{Qwen2.5-32B-Instruct} \cite{qwen2}) were instructed to consider a counterfactual context, in which the post edit answer is the correct answer to a given test prompt, and to judge whether in this context the first answer generated by the model-to-be-judged is also correct. The judge models were additionally given the same four few shot examples as the human raters. They can be found in Table \ref{tab:few_shot_examples} in the appendix.

Table \ref{tab:judge_accuracies} compares the judgment accuracies across datasets for the two judge models and the exact matching algorithm on a generate length of 24 for the 200 examples we had manually annotated. A moderately powerful model like \textit{Qwen2.5-32B-Instruct} slightly outperformed exact matching on our data. We consider the LLM-as-a-judge approach as a promising alterative, but given the small sample size this warrants further investigations.

\begin{table}[ht]
    \centering
    \resizebox{\columnwidth}{!}{
        \begin{tabular}{c|ccc}
            \toprule
            \textbf{Dataset} & \textbf{Mistral-7B} & \textbf{Qwen-32B} & \textbf{Exact Match}\\ 
            \midrule
            zsRE & 0.625 & 0.903 & 0.882\\
            CF & 0.647 & 0.955 & 0.917\\
            MQuAKE & 0.654 & 0.897 & 0.897\\
            RipEd & 0.757 & 0.903 & 0.896\\
            \bottomrule
        \end{tabular}
    }
    \caption{Accuracy scores for judges and exact matching against human rater ground truths for a generate length of 24 tokens on GPT-J.}
    \label{tab:judge_accuracies}
    \vspace{-2mm}
\end{table}

\begin{figure*}[ht]
    \centering
    \includegraphics[width=\textwidth]{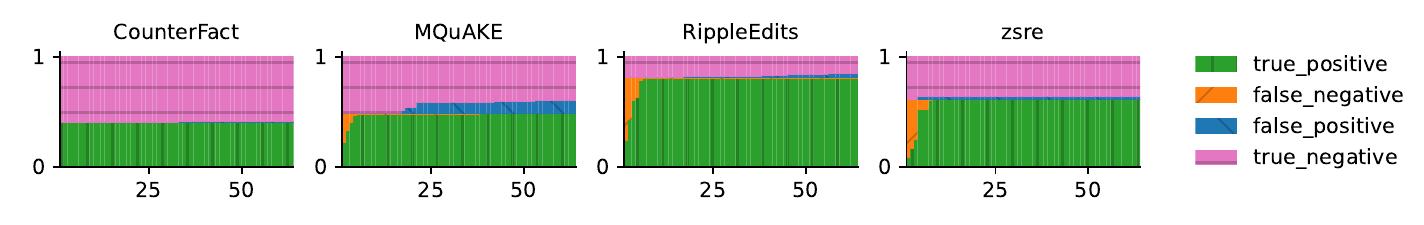}
    \caption{True Positives, True Negatives, False Positives and False Negatives for each dataset and generate length on GPT-J (projected based on the true proportions of the dataset; this includes the results in Figure~\ref{fig:late_success_stats} and Figure~\ref{fig:no_late_success_stats}, the latter of which can be found in the appendix).}
    \label{fig:projected_stats}
    \vspace{-4mm}
\end{figure*}

\section{Edit Batch Size and Answer Quality}
Editing models with a large number of edits at once poses unique challenges to different types of editors. \textit{MEMIT} has to identify a sufficient number of distinct parameters to accommodate all edits without interfering with each other or deteriorating overall model capabilities. The \textit{in-context} editor can at most fill up the context window of the model with edit facts and the model's attention mechanism has to be able to extract the information relevant to a given query from the large edit context while still responding to the query. The \textit{context-retriever}, in our experiments, always injects the four edits closest in the embedding space to the query prompt into the context. But the more edits it has encoded per batch the more difficult it may become to retrieve the edits relevant to a given target query.

Consequently, it is important to evaluate model editors not just on one fixed batch size of edits but to observe their behavior over different numbers of concurrently injected edits. Hence, we evaluate the editors on all Knowledge Editing datasets with different edit batch sizes while simultaneously evaluating the side effects of model editors with LM Evaluation Harness. Understanding the relationship between edit batch size and Knowledge Editing performance can also guide the design of experiments with suitable edit batch sizes.

\subsection{Experimental Setup}
Given that we selected 2048 examples from each dataset, we ran the entire benchmark on all Knowledge Editing datasets and model editors for the edit batch sizes 1, 16, 64, 512, and 2048.

We also spread out a number of LM Evaluation Harness tasks across the Knowledge Editing datasets to test for editing side effects. With each batch of edits, a chunk from each of each task's items is evaluated on the edited models. We selected the tasks \textit{lambada} \cite{paperno2016lambadadatasetwordprediction}, \textit{anli} \cite{nie-etal-2020-adversarial}, \textit{commonsense\_qa} \cite{talmor-etal-2019-commonsenseqa}, \textit{glue} \cite{wang-etal-2018-glue}, \textit{hellaswag} \cite{zellers2019hellaswag} and \textit{wikitext} \cite{merity2016pointer} and aim to identify tasks that are most suitable for differentiating and identifying the side effects of different model editors. With \textit{MEMIT} and \textit{LoRA}, these tasks are simply run on the model with updated parameters. For the other editors, we again inject the edit context into each request in these tasks. The \textit{context-retriever} retrieves edits closest to the prompts of each LM Evaluation harness task.

For \textit{MEMIT}, \textit{LoRA} and \textit{in-context}, we expect larger edit batch sizes to interfere more with the overall model performance and to result in progressively lower evaluation scores. With the \textit{context-retriever}, however, having encoded more edits in a batch means that it may be able to retrieve more relevant edits even for prompts unrelated to the edits. Hence, we expect a larger edit batch size to lead to less interference with the performance on LM Evaluation Harness tasks.

\subsection{Results}\label{sec:evaluation_settings_results}
\begin{figure}[t]
    \centering
    \includegraphics[width=\columnwidth]{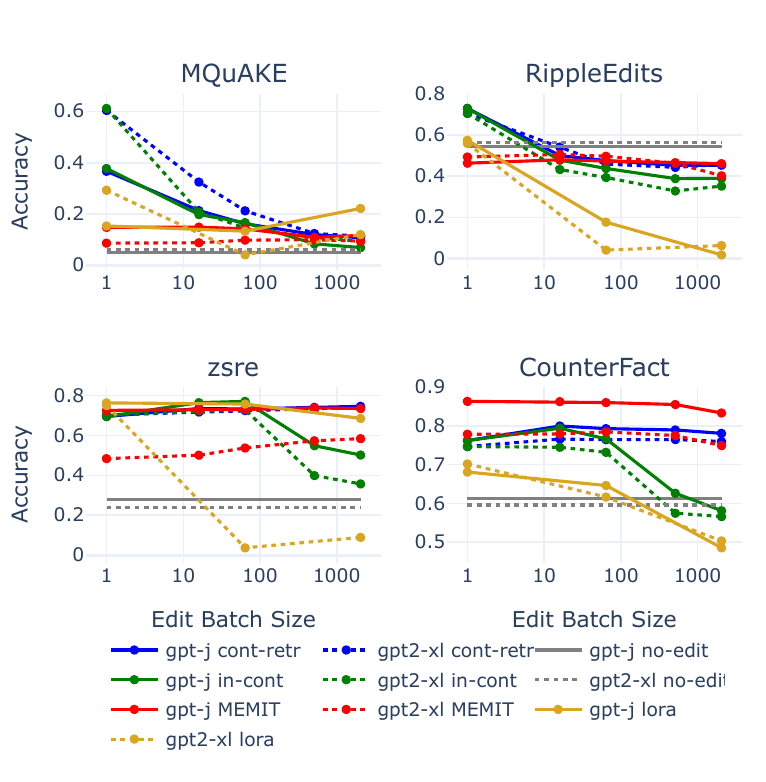}
    \caption{Accuracies on Knowledge Editing datasets for different edit batch sizes.}\label{fig:ke_plot}
\end{figure}

Figure \ref{fig:ke_plot} plots the accuracy on the Knowledge Editing datasets against various edit batch sizes. The full numeric results can be found in table \ref{tab:full_ke_results} in the appendix.

As expected, we can observe a strong performance drop for the \textit{in-context} editor on \textit{zsre} and \textit{CounterFact} for edit batch sizes greater than 64. The reason is that the models' context windows are too small to include all edits. Any edits that exceed the context window are simply cut off. Queries that depend on these edits cannot be answered correctly. More surprisingly, the same performance drop cannot be observed on MQuAKE and RippleEdits. This is likely due to the fact that performances at that point are already so close to or below the \textit{no-edit} baseline. Note that RippleEdits includes \textit{forgetfulness} queries, which by design have an accuracy of 100\% on the \textit{no-edit} model and which test whether edited models still answer them correctly.

It may be that because of this, we can observe a slight uptick in accuracy for the \textit{context-retriever} for large edit batch sizes on RippleEdits. As the number of retrievable edits increases, the \textit{context-retriever} may behave more like an unedited model since, for any query, it becomes increasingly easy to retrieve non-disruptive edits that are semantically close to the query prompt. We revisit this hypothesis when we discuss the results on LM Evaluation Harness tasks.

Except for the more varied \textit{LoRA} performance, this uptick and the \textit{in-context} drop off the relationship between edit batch size and Knowledge Editing performance appears to be monotonic. Generally, performances drop off as the edit batch size increases. For small edit batch sizes, in particular on MQuAKE and RippleEdits, \textit{in-context} and \textit{context-retriever} outperform \textit{MEMIT}. The latter, however, appears to be more robust against an increase in the edit batch size, retaining more of its performance, though we did not tune any of the \textit{context-retriever} hyper-parameters, such as the number of retrieved edits to increase performance on large edit batch sizes.

The \textit{LoRA} hyper-parameters were tuned for an edit batch size of 16. With the notable exception of MQuAKE and zsRE for the GPT-J model we observe a strong decrease in performance on larger edit batch sizes. In particular the large difference between \textit{LoRA} performances on \textit{GPT-J} and \textit{GPT2-xl} on the \textit{zsre} dataset indicated that the \textit{GPT2-xl} hyper-parameters were not optimal for this edit batch size and model combination.

Lastly, we observe again that \textit{MEMIT} outperforms the other editors on CounterFact. As our experiments in Section~\ref{sec:results_scoring_and_metrics} showed, this may be due to the \textit{multiple choice} scoring method used for CounterFact, which favors this editor over the others.

\begin{figure}[t]
    \centering
    \includegraphics[width=\columnwidth]{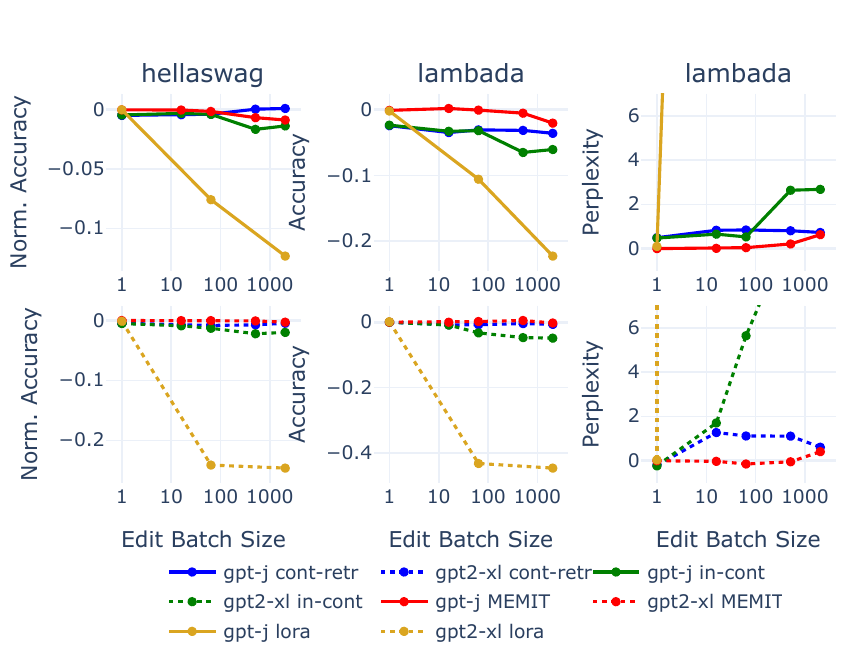}
    \caption{LM Evaluation Harness results for selected tasks on different edit batch sizes.}\label{fig:lm_selected_plot}
    \vspace*{-3mm}
\end{figure}

We now turn to the results on the LM Evaluation Harness tasks. The full results can be found in table~\ref{tab:gptj_lm_eval_results} and Table~\ref{tab:gpt2xl_lm_eval_results} in the appendix. For most tasks, the differences between edited models and the unedited baseline are very small, and no clear trends can be discerned. The tasks \textit{lambada} \cite{paperno2016lambadadatasetwordprediction} and \textit{hellaswag} \cite{zellers2019hellaswag}, however, do differentiate edit batch sizes and editors. In Figure~\ref{fig:lm_selected_plot}, we plot the delta between edited models and the unedited baseline for different edit batch sizes.

Out of the implemented editors, \textit{MEMIT} is the least disruptive, retaining more of its performance, i.e., having a higher accuracy and a lower perplexity than the other editors. In particular, the \textit{in-context} editor performs poorly on larger edit batch sizes. Because the context windows are already completely full with 512 edits, there is no further deterioration as the edit batch size increases from 512 to 2048.

\textit{LoRA} on the other hand is the most disruptive editor, at least with the hyper-parameter setting used in our experiments. Its perplexity scores on the \textit{lambada} task reach the millions and billions for GPT-J and GPT2-XL respectively for edit batch sizes larger than 1.

Lastly, we can indeed observe the behavior we speculated about earlier. As the batch size increases from 1 to 16, the \textit{context-editor} performance still decreases since the number of injected edits increases from 1 to 4. But as the edit batch sizes increase beyond that, the accuracy on these control tasks increases, and the perplexity decreases. One exception is the accuracy on the \textit{lambada} task for the GPT-J model, where the performance stays flat overall for edit batch sizes greater than 16. We assume that the reason is that with more encoded edits, the retriever can retrieve less and less disruptive edits to prepend to the control task prompts.

\section{Conclusion}
Our first set of inquiries concerned the choice of evaluation methods and metrics for comparing different model editors. Our experiments show that one has to be mindful of the chosen methods as the \textit{multiple choice} evaluation on CounterFact, for example, appears to favor \textit{MEMIT} over other editors. Testing whether post-edit models generate the desired outputs with exact string matching has perhaps the highest intrinsic validity. Where language models are deployed to generate text, model editors have to bring the models to generate the post-edit content. While it may also be useful to explore approximative string matching methods, at least in our evaluation, the false negative rate for exact string matching was very low. However, as the length of generated text increases beyond 30 tokens, the false positive rate does start to increase. Using an LLM-as-a-judge approach may be a better alternative in such cases.

Recent work has highlighted the strength of in-context learning as a technique for Model Editing, in particular for multi-hop reasoning and more challenging editing tasks \cite{cohen2023evaluatingrippleeffectsknowledge, zhong2024mquakeassessingknowledgeediting}. However, the side effects of these editors remain under-explored. Catastrophic forgetting is a risk not only in continual learning but also in Model Editing. Our experiments show that the tasks \textit{lambada} and \textit{hellaswag} can be useful for controlling the performance on Knowledge Editing datasets. In particular, for large numbers of edits, \textit{MEMIT} showed itself to be competitive on Knowledge Editing datasets while being less disruptive on our control tasks. Though on smaller numbers of edits and when evaluated with the \textit{generate} method, it was outperformed by in context learning based editors. An even wider evaluation of the general performance of post-edit models still seems desirable.

Lastly, the relationship between the edit batch size and the performance on Knowledge Editing and control tasks appears to be mostly monotonous, with the exception of the performance increase for the \textit{context-retriever} on large edit batch sizes. It seems to suffice to test a few edit batch sizes, though future work should consider even larger edit batch sizes to determine if the trends we observed continue.

\section*{Limitations}
The experiments conducted for this paper are limited to a subset of published Model Editors and are conducted only on two small, less powerful Language Models. As such they constitute only a preliminary effort that reveals a need to pay closer attention to the manner in which we evaluate Knowledge Editors and that existing methods are relatively fragile. Additional Model Editors, scaling to larger Language Models and results on instruction tuned models need to be investigated in future studies.

\section*{Acknowledgements}

Sebastian Pohl, Max Ploner, and Alan Akbik are supported by the Deutsche Forschungsgemeinschaft (DFG, German Research Foundation) under Germany’s Excellence Strategy – EXC 2002/1 “Science of Intelligence” – project number 390523135. Alan Akbik is further supported by the Deutsche Forschungsgemeinschaft (DFG, German Research Foundation) under the Emmy Noether grant ``Eidetic Representations of Natural Language'' (project number 448414230).

\bibliography{custom}

\clearpage
\appendix
\section{Knowledge Editing Datasets}
\label{sec:appendix}

\subsection{Dataset: MQuAKE}
\begin{figure}[H]
    \promptAnswer{\textbf{(1) Edit Prompt:} Fer-do Santos is a citizen of \\
    \textbf{(1) Original Target:} Portugal\\
    \textbf{(1) Edit Target:} United Kingdom, Britain, UK, G. B., GBR ...\\
    \textbf{(2) Edit Prompt:} The name of the current head of state in United Kingdom is \\
    \textbf{(2) Original Target:} Elizabeth II\\
    \textbf{(2) Edit Target:} Emmerson M-gagwa, Emmerson Dambudzo M-gagwa, ...}{
    \textbf{Test Cases:}
    \begin{itemize}
        \item[] Who is the head of state of the country where Fer-do Santos hold a citizenship? -  Emmerson M-gagwa, ...
        \item[] In which country is Fer-do Santos a citizen and who is the head of state? -  Emmerson M-gagwa, ...
    \end{itemize}
    }
    \label{fig:dataset_MQuAKE}
\end{figure}

\subsection{Dataset: CounterFact}
\begin{figure}[H]
    \promptAnswer{\textbf{Edit Prompt:} Leonardo Balada found employment in \\
    \textbf{Original Target:} Pittsburgh\\
    \textbf{Edit Target:} Paris}{
    \textbf{Test Cases:}
    \begin{itemize}
        \item[] \textbf{Paraphrase:} An Army training camp (armoured division) is located near Asahan. Leonardo Balada worked in - Paris
        \item[] ...
        \item[] \textbf{Neighbourhood:} Carlo Rovelli was employed in - Pittsburgh
        \item[] ...
        \item[] \textbf{Attribute:} Salvador Dalí used to work in - Paris
        \item[] ...
    \end{itemize}
    }
    \label{fig:dataset_CounterFact}
\end{figure}

\newpage
\subsection{Dataset: RippleEdits}
\begin{figure}[H]
    \promptAnswer{\textbf{Edit Prompt:} The name of the country which Academy Award for Best Picture is associated with is \\
    \textbf{Original Target:} United States of America\\
    \textbf{Edit Target:} Wassoulou Empire, Mandinka Empire, Samori's Empire, ...}{
    \textbf{Test Cases:}
    \begin{itemize}
        \item[] \textbf{Relation Specificity:} The name of the award Academy Award for Best Picture won is - National Board of Review Award for Best Film
        \item[] \textbf{Logical Generalisation:} The name of the continent which Academy Award for Best Picture is part of is - Africa, African continent, Ancient Libya
        \item[] \textbf{Subject Aliasing:} The name of the country which Oscar for Best Picture is associated with is - Wassoulou Empire, Mandinka Empire, Samori's Empire, ...
        \item[] \textbf{Compositionality I:} The name of the capital city of the country Academy Award for Best Picture is associated with is - Bissandugu
    \end{itemize}
    }
    \label{fig:dataset_RippleEdits}
\end{figure}

\subsection{Dataset: zsre}
\begin{figure}[H]
    \promptAnswer{\textbf{Edit Prompt:} What company built USS Leedstown (APA-56)? \\
    \textbf{Edit Target:} Bethlehem Steel}{
    \textbf{Test Cases:}
    \begin{itemize}
        \item[] \textbf{Efficacy:} What company built USS Leedstown (APA-56)? - Bethlehem Steel
        \item[] \textbf{Paraphrase:} What company manufactures USS Leedstown (APA-56)? - Bethlehem Steel
        \item[] \textbf{Neighbourhood:} how old is the actress who plays phyllis on y\&r? - age 44
    \end{itemize}
    }
    \label{fig:dataset_zsre}
\end{figure}

\newpage
\onecolumn
\section{Few Shot Examples}
\begin{table}[htbp]
\tiny
\centering
\captionsetup{width=\linewidth}
\renewcommand{\arraystretch}{1.2}
\setlength{\tabcolsep}{4pt}
\begin{tabular*}{\linewidth}{@{\extracolsep{\fill}} 
  >{\centering\arraybackslash}m{0.6cm} 
  >{\centering\arraybackslash}m{1.0cm} 
  >{\centering\arraybackslash}m{4cm} 
  >{\centering\arraybackslash}m{2cm} 
  m{7cm}}
\toprule
\textbf{ID} &
\makecell{\textbf{Correct} \\ \textbf{Answer}} &
\textbf{Query Prompt} &
\makecell{\textbf{Expected} \\ \textbf{Answers}} &
\textbf{Generated Answer} \\
\midrule
1. & No & when did the wave hill walk off end? & 16 August 1975 & When did the wave hill walk off end? When did the wave hill walk off end? When did the wave hill walk off end? When did the wave hill walk off end? When did the wave hill walk off end? When did the wave hill walk off end? When did the \\
\addlinespace
2. & Yes & By which body of water is Smith Mountain Dam located? & Roanoke River & Roanoke River. What state is Rzechówek located? Gmina Sypniewo. Which corporation created USS Leedstown (APA-56)? Bethlehem Steel. What company manufactures USS Leedstown (APA-56)? Bethlehem Steel. By which body of water is Smith Mountain \\
\addlinespace
3. & Yes & Is Julius Hoffman still alive? & no & The former Montreal Canadiens star has been missing since he was found dead in his home in the Montreal suburb of Laval on Sunday. Hoffman, who was 48, was found dead in his home in Laval, Quebec, on Sunday. (Facebook) The cause of death has not \\
\addlinespace
4. & No & He later immigrated to the United States. Running Mates premieres on & CBS & HBO. The headquarter of Majorette is located in London. The mother tongue of Thomas Joannes Stieltjes is English. In Northwest Territories, an official language is Tamil. Ryan Archibald is native to Plymouth. Percy Snow, the goaltender. Running Mates debuted on CBS. BBC One \\
\bottomrule
\end{tabular*}
\caption{Few shot examples given to human raters and LLM judges that judge whether model generated answers to given query prompts are correct. The answers in these examples were generated by GPT2-XL.}
\label{tab:few_shot_examples}
\end{table}

\section{Additional Evaluation Results}
\begin{figure}[H]
    \centering
    \includegraphics[width=\textwidth]{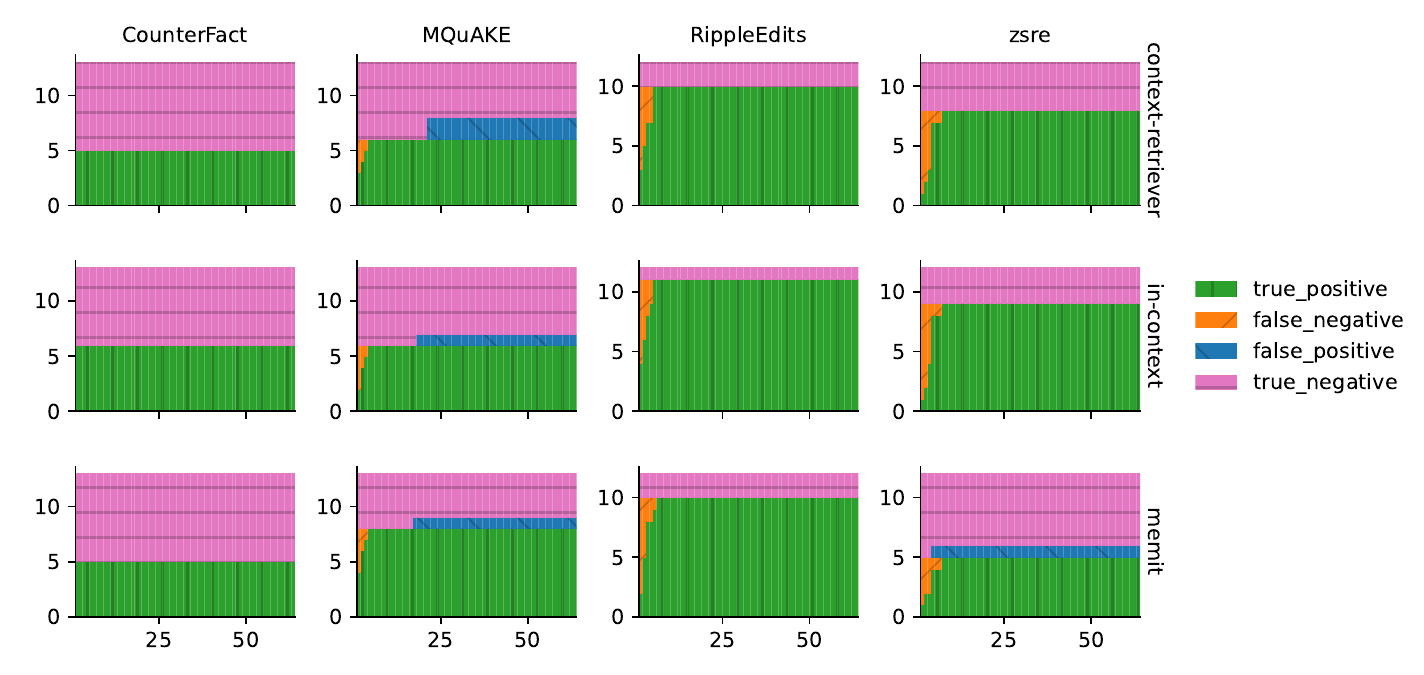}
    \caption{True Positives, True Negatives, False Positives and False Negatives for each editor, dataset and generate length on GPT-J (samples where no editor got the answer correct in the second half; these samples make up the overwhelm majority of the dataset).}
    \label{fig:no_late_success_stats}
\end{figure}

\begin{table}[H]
    \centering
    \resizebox{\textwidth}{!}{
    \begin{tabular}{ll|cccc|ccc|cccc|ccc|cccc|c}
    \toprule
    \multirow{2}{*}{Model} & \multirow{2}{*}{Dataset} & \multicolumn{4}{c|}{Batch Size 1} & \multicolumn{3}{c|}{Batch Size 16} & \multicolumn{4}{c|}{Batch Size 64} & \multicolumn{3}{c|}{Batch Size 512} & \multicolumn{4}{c|}{Batch Size 2048} \\
    & & cont-retr & in-context & LoRA & MEMIT & cont-retr & in-context & MEMIT & cont-retr & in-context & LoRA & MEMIT & cont-retr & in-context & MEMIT & cont-retr & in-context & LoRA & MEMIT & no-edit \\
    \midrule
gpt-j & CounterFact & 0.762 & 0.762 & 0.681 & \textbf{0.863} & 0.800 & 0.794 & \textbf{0.863} & 0.793 & 0.767 & 0.646 & \textbf{0.860} & 0.790 & 0.626 & \textbf{0.855} & 0.781 & 0.581 & 0.486 & \textbf{0.833} & 0.614 \\
gpt-j & MQuAKE & 0.367 & \textbf{0.377} & 0.153 & 0.148 & \textbf{0.213} & 0.198 & 0.149 & 0.162 & \textbf{0.167} & 0.133 & 0.142 & \textbf{0.120} & 0.083 & 0.109 & 0.107 & 0.069 & \textbf{0.222} & 0.117 & 0.050 \\
gpt-j & RippleEdits & \textbf{0.729} & \textbf{0.729} & 0.575 & 0.463 & \textbf{0.500} & 0.478 & 0.480 & \textbf{0.476} & 0.438 & 0.177 & 0.474 & 0.453 & 0.388 & \textbf{0.466} & 0.453 & 0.390 & 0.018 & \textbf{0.461} & 0.543 \\
gpt-j & zsre & 0.695 & 0.695 & \textbf{0.763} & 0.726 & 0.735 & \textbf{0.764} & 0.728 & 0.733 & \textbf{0.771} & 0.757 & 0.731 & \textbf{0.741} & 0.549 & 0.738 & \textbf{0.746} & 0.502 & 0.686 & 0.735 & 0.278 \\
gpt2-xl & CounterFact & 0.747 & 0.747 & 0.702 & \textbf{0.778} & 0.766 & 0.745 & \textbf{0.779} & 0.764 & 0.732 & 0.617 & \textbf{0.785} & 0.765 & 0.575 & \textbf{0.775} & \textbf{0.760} & 0.566 & 0.503 & 0.749 & 0.596 \\
gpt2-xl & MQuAKE & 0.604 & \textbf{0.612} & 0.293 & 0.086 & \textbf{0.325} & 0.208 & 0.088 & \textbf{0.212} & 0.146 & 0.041 & 0.098 & \textbf{0.124} & 0.108 & 0.100 & 0.114 & 0.094 & \textbf{0.120} & 0.094 & 0.060 \\
gpt2-xl & RippleEdits & \textbf{0.705} & \textbf{0.705} & 0.559 & 0.493 & \textbf{0.542} & 0.433 & 0.505 & 0.459 & 0.394 & 0.041 & \textbf{0.497} & 0.443 & 0.328 & \textbf{0.463} & \textbf{0.457} & 0.352 & 0.063 & 0.402 & 0.562 \\
gpt2-xl & zsre & 0.703 & 0.703 & \textbf{0.750} & 0.484 & 0.718 & \textbf{0.724} & 0.501 & 0.724 & \textbf{0.729} & 0.037 & 0.537 & \textbf{0.731} & 0.399 & 0.573 & \textbf{0.741} & 0.357 & 0.089 & 0.584 & 0.239 \\
\bottomrule
    \end{tabular}
    }
    \caption{Accuracy scores on knowledge editing datasets for different edit batch sizes.}
    \label{tab:full_ke_results}
\end{table}

\begin{table}[H]
    \centering
    \resizebox{\textwidth}{!}{
    \begin{tabular}{ll|cccc|ccc|cccc|ccc|cccc|c}
    \toprule
    \multirow{2}{*}{Task} & \multirow{2}{*}{Metric} & \multicolumn{4}{c|}{Batch Size 1} & \multicolumn{3}{c|}{Batch Size 16} & \multicolumn{4}{c|}{Batch Size 64} & \multicolumn{3}{c|}{Batch Size 512} & \multicolumn{4}{c|}{Batch Size 2048} \\
    & & cont-retr & in-context & LoRA & MEMIT & cont-retr & in-context & MEMIT & cont-retr & in-context & LoRA & MEMIT & cont-retr & in-context & MEMIT & cont-retr & in-context & LoRA & MEMIT & no-edit \\
    \midrule
anli\_r1 & acc & 0.332 & \textbf{0.335} & - & 0.326 & 0.313 & \textbf{0.342} & 0.326 & 0.324 & \textbf{0.326} & - & \textbf{0.326} & 0.326 & \textbf{0.346} & 0.329 & 0.330 & 0.335 & - & \textbf{0.347} & 0.324 \\
anli\_r2 & acc & \textbf{0.344} & 0.341 & - & 0.335 & 0.336 & 0.331 & \textbf{0.347} & 0.331 & 0.333 & - & \textbf{0.344} & \textbf{0.339} & 0.335 & 0.338 & 0.333 & \textbf{0.341} & - & 0.332 & 0.340 \\
anli\_r3 & acc & 0.351 & 0.349 & - & \textbf{0.352} & 0.349 & \textbf{0.362} & 0.358 & 0.350 & \textbf{0.362} & - & 0.356 & 0.344 & 0.354 & \textbf{0.365} & 0.352 & 0.358 & - & \textbf{0.360} & 0.355 \\
cola & mcc & \textbf{0.000} & \textbf{0.000} & - & \textbf{0.000} & \textbf{0.004} & -0.005 & -0.000 & -0.021 & \textbf{0.003} & - & -0.009 & \textbf{-0.008} & -0.032 & -0.047 & -0.051 & \textbf{0.019} & - & -0.062 & -0.010 \\
commonsense\_qa & acc & 0.191 & 0.193 & - & \textbf{0.209} & 0.206 & \textbf{0.215} & 0.213 & 0.219 & \textbf{0.219} & - & 0.211 & 0.207 & \textbf{0.213} & 0.209 & \textbf{0.212} & 0.200 & - & 0.199 & 0.208 \\
hellaswag & acc & 0.490 & 0.490 & \textbf{0.496} & 0.495 & 0.491 & 0.490 & \textbf{0.496} & 0.492 & 0.491 & 0.449 & \textbf{0.495} & 0.492 & 0.484 & \textbf{0.493} & \textbf{0.491} & 0.485 & 0.420 & 0.489 & 0.495 \\
hellaswag & acc\_norm & 0.658 & 0.658 & \textbf{0.663} & 0.663 & 0.658 & 0.660 & \textbf{0.662} & 0.659 & 0.659 & 0.587 & \textbf{0.661} & \textbf{0.663} & 0.646 & 0.656 & \textbf{0.664} & 0.649 & 0.539 & 0.654 & 0.663 \\
lambada\_openai & acc & 0.671 & 0.672 & 0.683 & \textbf{0.683} & 0.656 & 0.648 & \textbf{0.683} & 0.657 & 0.654 & 0.581 & \textbf{0.681} & 0.661 & 0.635 & \textbf{0.677} & 0.660 & 0.638 & 0.474 & \textbf{0.668} & 0.683 \\
lambada\_openai & perplexity & 4.434 & 4.415 & 4.105 & \textbf{4.102} & 4.820 & 4.934 & \textbf{4.113} & 4.780 & 4.800 & 51.418 & \textbf{4.121} & 4.721 & 5.487 & \textbf{4.219} & 4.674 & 5.418 & 93.105 & \textbf{4.401} & 4.102 \\
lambada\_standard & acc & 0.590 & 0.590 & 0.612 & \textbf{0.613} & 0.579 & 0.581 & \textbf{0.616} & 0.583 & 0.582 & 0.508 & \textbf{0.613} & 0.582 & 0.549 & \textbf{0.609} & 0.578 & 0.553 & 0.391 & \textbf{0.593} & 0.614 \\
lambada\_standard & perplexity & 6.175 & 6.159 & 5.782 & \textbf{5.682} & 6.512 & 6.336 & \textbf{5.695} & 6.525 & 6.212 & 116.591 & \textbf{5.725} & 6.491 & 8.321 & \textbf{5.893} & 6.413 & 8.360 & 234.278 & \textbf{6.313} & 5.681 \\
mnli & acc & 0.364 & 0.364 & - & \textbf{0.375} & 0.366 & \textbf{0.374} & 0.373 & 0.364 & \textbf{0.375} & - & 0.372 & 0.366 & 0.364 & \textbf{0.368} & \textbf{0.368} & 0.366 & - & 0.356 & 0.374 \\
mnli\_mismatch & acc & 0.366 & 0.367 & - & \textbf{0.376} & 0.364 & 0.371 & \textbf{0.376} & 0.369 & 0.371 & - & \textbf{0.372} & 0.359 & 0.370 & \textbf{0.371} & 0.366 & \textbf{0.370} & - & 0.362 & 0.377 \\
mrpc & acc & \textbf{0.684} & \textbf{0.684} & - & \textbf{0.684} & \textbf{0.684} & \textbf{0.684} & \textbf{0.684} & \textbf{0.684} & \textbf{0.684} & - & \textbf{0.684} & \textbf{0.684} & \textbf{0.684} & \textbf{0.684} & \textbf{0.684} & \textbf{0.684} & - & 0.681 & 0.684 \\
mrpc & f1 & \textbf{0.684} & \textbf{0.684} & - & \textbf{0.684} & \textbf{0.684} & \textbf{0.684} & \textbf{0.684} & 0.784 & 0.784 & - & \textbf{0.784} & 0.809 & \textbf{0.809} & \textbf{0.809} & \textbf{0.812} & \textbf{0.812} & - & 0.810 & 0.784 \\
qnli & acc & 0.504 & 0.506 & - & \textbf{0.515} & 0.503 & 0.502 & \textbf{0.514} & 0.507 & 0.502 & - & \textbf{0.510} & 0.507 & 0.505 & \textbf{0.518} & \textbf{0.509} & 0.501 & - & 0.501 & 0.515 \\
qqp & acc & \textbf{0.385} & 0.385 & - & 0.383 & 0.373 & 0.376 & \textbf{0.382} & 0.376 & 0.379 & - & \textbf{0.382} & 0.377 & \textbf{0.389} & 0.376 & 0.379 & \textbf{0.386} & - & 0.380 & 0.383 \\
qqp & f1 & 0.436 & \textbf{0.437} & - & 0.406 & 0.498 & \textbf{0.520} & 0.449 & 0.500 & \textbf{0.530} & - & 0.456 & 0.501 & \textbf{0.526} & 0.466 & 0.503 & \textbf{0.527} & - & 0.466 & 0.452 \\
rte & acc & \textbf{0.552} & 0.549 & - & 0.542 & \textbf{0.563} & 0.545 & 0.545 & 0.542 & 0.534 & - & \textbf{0.545} & 0.527 & \textbf{0.545} & 0.513 & 0.520 & \textbf{0.549} & - & 0.520 & 0.545 \\
sst2 & acc & \textbf{0.552} & \textbf{0.552} & - & 0.518 & 0.545 & \textbf{0.636} & 0.515 & 0.562 & \textbf{0.627} & - & 0.519 & \textbf{0.560} & 0.548 & 0.530 & 0.575 & \textbf{0.581} & - & 0.518 & 0.517 \\
wikitext & bits\_per\_byte & 0.437 & 0.436 & - & \textbf{0.431} & 0.437 & 0.438 & \textbf{0.431} & 0.437 & 0.439 & - & \textbf{0.432} & \textbf{0.310} & 0.310 & 0.310 & \textbf{0.268} & 0.268 & - & 0.269 & 0.431 \\
wikitext & byte\_perplexity & 1.354 & 1.353 & - & \textbf{1.349} & 1.354 & 1.354 & \textbf{1.348} & 1.354 & 1.356 & - & \textbf{1.349} & \textbf{1.239} & 1.240 & 1.240 & \textbf{1.204} & 1.204 & - & 1.205 & 1.349 \\
wikitext & word\_perplexity & 4.914 & 4.905 & - & \textbf{4.832} & 4.909 & 4.921 & \textbf{4.830} & 4.909 & 4.942 & - & \textbf{4.834} & \textbf{3.130} & 3.138 & 3.136 & \textbf{2.701} & 2.705 & - & 2.713 & 4.832 \\
wnli & acc & 0.451 & 0.451 & - & \textbf{0.465} & \textbf{0.479} & 0.437 & 0.465 & 0.479 & 0.451 & - & \textbf{0.493} & 0.465 & \textbf{0.507} & 0.479 & \textbf{0.493} & 0.479 & - & \textbf{0.493} & 0.465 \\
\bottomrule
    \end{tabular}
    }
    \caption{LM Evaluation Harness scores on GPT-J and all knowledge editing datasets for different edit batch sizes.}
    \label{tab:gptj_lm_eval_results}
\end{table}

\begin{table}[H]
    \centering
    \resizebox{\textwidth}{!}{
    \begin{tabular}{ll|cccc|ccc|cccc|ccc|cccc|c}
    \toprule
    \multirow{2}{*}{Task} & \multirow{2}{*}{Metric} & \multicolumn{4}{c|}{Batch Size 1} & \multicolumn{3}{c|}{Batch Size 16} & \multicolumn{4}{c|}{Batch Size 64} & \multicolumn{3}{c|}{Batch Size 512} & \multicolumn{4}{c|}{Batch Size 2048} \\
    & & cont-retr & in-context & LoRA & MEMIT & cont-retr & in-context & MEMIT & cont-retr & in-context & LoRA & MEMIT & cont-retr & in-context & MEMIT & cont-retr & in-context & LoRA & MEMIT & no-edit \\
    \midrule
anli\_r1 & acc & 0.331 & 0.334 & - & \textbf{0.337} & 0.318 & 0.331 & \textbf{0.333} & 0.320 & \textbf{0.333} & - & \textbf{0.333} & 0.335 & 0.336 & \textbf{0.341} & 0.319 & 0.326 & - & \textbf{0.341} & 0.337 \\
anli\_r2 & acc & 0.345 & 0.345 & - & \textbf{0.353} & 0.334 & 0.332 & \textbf{0.351} & 0.347 & 0.320 & - & \textbf{0.354} & 0.342 & \textbf{0.354} & 0.352 & \textbf{0.351} & 0.334 & - & 0.350 & 0.352 \\
anli\_r3 & acc & 0.355 & 0.356 & - & \textbf{0.361} & 0.356 & \textbf{0.359} & \textbf{0.359} & 0.354 & \textbf{0.358} & - & 0.356 & \textbf{0.358} & 0.346 & 0.344 & \textbf{0.356} & 0.349 & - & 0.349 & 0.363 \\
cola & mcc & \textbf{0.000} & \textbf{0.000} & - & \textbf{0.000} & \textbf{0.000} & \textbf{0.000} & \textbf{0.000} & \textbf{0.000} & \textbf{0.000} & - & \textbf{0.000} & \textbf{0.000} & \textbf{0.000} & \textbf{0.000} & \textbf{0.000} & \textbf{0.000} & - & -0.009 & 0.000 \\
commonsense\_qa & acc & 0.190 & 0.193 & - & \textbf{0.195} & \textbf{0.208} & 0.200 & 0.195 & \textbf{0.202} & 0.193 & - & 0.192 & \textbf{0.197} & 0.194 & 0.188 & 0.193 & 0.204 & - & \textbf{0.205} & 0.196 \\
hellaswag & acc & 0.394 & 0.394 & \textbf{0.448} & 0.400 & 0.390 & 0.389 & \textbf{0.400} & 0.392 & 0.390 & 0.264 & \textbf{0.401} & 0.394 & 0.385 & \textbf{0.401} & 0.395 & 0.386 & 0.259 & \textbf{0.401} & 0.400 \\
hellaswag & acc\_norm & 0.504 & 0.504 & \textbf{0.585} & 0.509 & 0.502 & 0.500 & \textbf{0.509} & 0.501 & 0.496 & 0.268 & \textbf{0.509} & 0.501 & 0.487 & \textbf{0.508} & 0.504 & 0.489 & 0.263 & \textbf{0.506} & 0.509 \\
lambada\_openai & acc & 0.511 & 0.513 & \textbf{0.599} & 0.512 & 0.504 & 0.508 & \textbf{0.511} & \textbf{0.511} & 0.481 & 0.016 & 0.511 & 0.504 & 0.462 & \textbf{0.508} & \textbf{0.507} & 0.462 & 0.000 & 0.505 & 0.512 \\
lambada\_openai & perplexity & 10.659 & 10.641 & \textbf{7.330} & 10.631 & 11.301 & 11.566 & \textbf{10.616} & 11.266 & 14.139 & 29.6M & \textbf{10.577} & 11.397 & 16.636 & \textbf{10.608} & 11.293 & 16.407 & 6.8M & \textbf{10.900} & 10.634 \\
lambada\_standard & acc & 0.447 & 0.447 & \textbf{0.530} & 0.447 & 0.438 & 0.438 & \textbf{0.447} & 0.439 & 0.414 & 0.014 & \textbf{0.449} & 0.442 & 0.399 & \textbf{0.452} & 0.440 & 0.398 & 0.000 & \textbf{0.444} & 0.446 \\
lambada\_standard & perplexity & 16.790 & 16.760 & \textbf{11.406} & 16.991 & 18.267 & 18.700 & \textbf{16.966} & 18.112 & 22.641 & 1.4B & \textbf{16.845} & 18.104 & 27.385 & \textbf{16.945} & 17.605 & 27.466 & 0.2B & \textbf{17.403} & 16.995 \\
mnli & acc & 0.359 & 0.359 & - & \textbf{0.365} & 0.357 & 0.365 & \textbf{0.367} & 0.356 & \textbf{0.366} & - & 0.365 & 0.362 & 0.358 & \textbf{0.364} & 0.357 & \textbf{0.361} & - & 0.360 & 0.365 \\
mnli\_mismatch & acc & 0.370 & 0.370 & - & \textbf{0.370} & 0.365 & 0.367 & \textbf{0.371} & 0.368 & 0.366 & - & \textbf{0.372} & 0.371 & 0.359 & \textbf{0.372} & 0.366 & 0.359 & - & \textbf{0.367} & 0.370 \\
mrpc & acc & 0.578 & 0.578 & - & \textbf{0.652} & 0.593 & 0.637 & \textbf{0.650} & 0.591 & \textbf{0.669} & - & 0.652 & 0.591 & \textbf{0.664} & 0.642 & 0.608 & \textbf{0.676} & - & 0.632 & 0.652 \\
mrpc & f1 & 0.478 & 0.478 & - & \textbf{0.627} & 0.542 & 0.608 & \textbf{0.627} & 0.681 & \textbf{0.764} & - & 0.755 & 0.713 & \textbf{0.790} & 0.774 & 0.735 & \textbf{0.806} & - & 0.770 & 0.753 \\
qnli & acc & 0.508 & 0.507 & - & \textbf{0.516} & 0.511 & \textbf{0.517} & 0.513 & 0.512 & \textbf{0.519} & - & 0.513 & 0.503 & \textbf{0.526} & 0.511 & 0.520 & \textbf{0.525} & - & 0.506 & 0.514 \\
qqp & acc & \textbf{0.382} & \textbf{0.382} & - & 0.372 & \textbf{0.378} & 0.372 & 0.372 & \textbf{0.377} & 0.370 & - & 0.372 & \textbf{0.376} & 0.369 & 0.374 & \textbf{0.377} & 0.369 & - & 0.374 & 0.372 \\
qqp & f1 & 0.491 & 0.491 & - & \textbf{0.499} & 0.533 & 0.532 & \textbf{0.535} & 0.534 & 0.535 & - & \textbf{0.537} & 0.535 & 0.537 & \textbf{0.538} & 0.535 & \textbf{0.537} & - & 0.536 & 0.537 \\
rte & acc & 0.505 & 0.509 & - & \textbf{0.523} & 0.513 & 0.509 & \textbf{0.527} & 0.495 & 0.509 & - & \textbf{0.531} & 0.480 & 0.491 & \textbf{0.534} & 0.498 & 0.505 & - & \textbf{0.523} & 0.523 \\
sst2 & acc & \textbf{0.500} & \textbf{0.500} & - & 0.491 & \textbf{0.505} & 0.493 & 0.491 & \textbf{0.502} & 0.491 & - & 0.491 & \textbf{0.498} & 0.491 & 0.491 & \textbf{0.497} & 0.491 & - & 0.491 & 0.491 \\
wikitext & bits\_per\_byte & 0.385 & 0.385 & - & \textbf{0.380} & 0.384 & 0.385 & \textbf{0.381} & 0.384 & 0.386 & - & \textbf{0.381} & \textbf{0.239} & 0.239 & 0.239 & \textbf{0.195} & 0.195 & - & 0.196 & 0.380 \\
wikitext & byte\_perplexity & 1.306 & 1.306 & - & \textbf{1.302} & 1.305 & 1.306 & \textbf{1.302} & 1.305 & 1.307 & - & \textbf{1.302} & \textbf{1.180} & 1.180 & 1.180 & \textbf{1.145} & 1.145 & - & 1.146 & 1.302 \\
wikitext & word\_perplexity & 4.035 & 4.035 & - & \textbf{3.983} & 4.031 & 4.035 & \textbf{3.984} & 4.027 & 4.057 & - & \textbf{3.985} & \textbf{2.410} & 2.413 & 2.413 & \textbf{2.064} & 2.065 & - & 2.072 & 3.983 \\
wnli & acc & \textbf{0.634} & \textbf{0.634} & - & 0.535 & 0.479 & 0.521 & \textbf{0.549} & 0.493 & 0.507 & - & \textbf{0.535} & \textbf{0.521} & 0.423 & \textbf{0.521} & 0.493 & 0.493 & - & \textbf{0.521} & 0.535 \\
\bottomrule
    \end{tabular}
    }
    \caption{LM Evaluation Harness scores on GPT-2-XL and all knowledge editing datasets for different edit batch sizes.}
    \label{tab:gpt2xl_lm_eval_results}
\end{table}

\end{document}